\documentclass[10pt,twocolumn,letterpaper]{article}

\usepackage{tabularx}
\usepackage{booktabs}
\usepackage{nicefrac}
\usepackage{multirow}
\usepackage{caption}
\usepackage{subfigure}
\usepackage{cvpr}              

\usepackage[dvipsnames]{xcolor}

\definecolor{cvprblue}{rgb}{0.21,0.49,0.74}
\usepackage[pagebackref,breaklinks,colorlinks,citecolor=cvprblue]{hyperref}

\def\paperID{3234} 
\def\confName{CVPR}
\def\confYear{2025}

\title{Frequency-Adaptive Low-Latency Object Detection Using Events and Frames}

\author{
Haitian Zhang$^{1, *}$ \quad Xiangyuan Wang$^{1, *}$ \quad  Chang Xu$^2$ \quad Xinya Wang$^1$ \\ Fang Xu$^1$ \quad Huai Yu$^1$ \quad Lei Yu$^1$  \quad Wen Yang$^{1, \dagger}$ \\
$^1$Wuhan University, Wuhan, China \quad $^2$EPFL, Lausanne, Switzerland\\
{\tt \small $^1$\{haitian.zhang, wangxiangyuan, xinya.wang, xufang, yuhuai,  ly.wd, yangwen\}@whu.edu.cn}\\
{\tt \small $^2$chang.xu@epfl.ch}\\
\url{https://github.com/Hatins/FAOD-master}
}

\begin{document}

\maketitle

\begin{abstract}
Fusing Events and RGB images for object detection leverages the robustness of Event cameras in adverse environments and the rich semantic information provided by RGB cameras. 
However, two critical mismatches: low-latency Events \textit{vs.}~high-latency RGB frames; temporally sparse labels in training \textit{vs.}~continuous flow in inference, significantly hinder the high-frequency fusion-based object detection.
To address these challenges, we propose the \textbf{F}requency-\textbf{A}daptive Low-Latency \textbf{O}bject \textbf{D}etector (FAOD). 
FAOD aligns low-frequency RGB frames with high-frequency Events through an Align Module, which reinforces cross-modal style and spatial proximity to address the Event-RGB Mismatch.
We further propose a training strategy, Time Shift, which enforces the module to align the prediction from temporally shifted Event-RGB pairs and their original representation, that is, consistent with Event-aligned annotations.
This strategy enables the network to use high-frequency Event data as the primary reference while treating low-frequency RGB images as supplementary information, retaining the low-latency nature of the Event stream toward high-frequency detection. Furthermore, we observe that these corrected Event-RGB pairs demonstrate better generalization from low training frequency to higher inference frequencies compared to using Event data alone.  
Extensive experiments on the PKU-DAVIS-SOD and DSEC-Detection datasets demonstrate that our FAOD achieves SOTA performance. Specifically, in the PKU-DAVIS-SOD Dataset, FAOD achieves 9.8 points improvement in terms of the mAP in fully paired Event-RGB data with only a quarter of the parameters compared to SODFormer, and even maintains robust performance (only a 3 points drop in mAP) under 80$\times$ Event-RGB frequency mismatch.


\end{abstract}

    
\section{Introduction}
Event cameras \cite{event_survey_2020_TPAMI}, with their sub-millisecond temporal resolution and high dynamic range (\textgreater 120 dB), offer the potential to tackle challenges such as detecting fast-moving objects and performing object detection in extreme lighting conditions \cite{event_dataset_2020_NIPS, event_detector_2023_CVPR}, \textit{i.e.}, overexposed and underexposed environments. However, compared to the RGB modality, Event can only provide sparse pulse information and lacks absolute intensity, resulting in inferior semantic information. 
Therefore, fusing both modalities provides the potential to simultaneously leverage the merits from both sides \cite{event_fusion_2023_TPAMI, JIF, cao2024embracing, huai_evlsd} to achieve higher accuracy and robustness.

Despite this potential, the fusion of Event and RGB can come at the sack of high temporal resolution \cite{cao2024embracing, event_fusion_2023_TPAMI}, a defining feature of Event stream \cite{event_detector_2023_CVPR, event_detector_2024_SAST}. Two critical challenges impeding the high-frequency fusion-based object detection can be summarized as Event-RGB Mismatch and Train-Infer Mismatch, as explained in Figure \ref{fig:first_fig}. 
Event-RGB Mismatch results from the different sampling frequencies of Event and RGB cameras \cite{event_fusion_2023_TPAMI, event_2023_tracking}, where the former can capture over $10,000$ frames (dense Event representation) per second while the latter typically can only capture around 60 frames \cite{event_survey_2020_TPAMI} per second. 
On the other hand, the Train-Infer Mismatch, first highlighted in work \cite{event_detector_2024_SSM}, refers to the inference frequency exceeding the training frequency, leading to a severe performance degradation \cite{event_detector_2023_CVPR}. Because of the substantial cost of acquiring high-frequency annotated data, models are typically trained on low-frequency annotations. This necessitates that models generalize effectively from low-frequency training to high-frequency inference.

To address all the challenges mentioned above, while maintaining high detection performance and fast inference speed, we design a robust Event-RGB fusion detection framework FAOD.
To address the Event-RGB Mismatch, we propose an Align Module that rectifies the RGB frames, ensuring the style and spatial proximity between RGB features and Event features. To activate the offset generator in the Align Module, we design a training strategy called Time Shift, which simulates the misalignment caused by frequency discrepancies between the two modalities and enforces consistency between the outputs with and without the introduced offsets, to ensure that the module is adequately trained. 
This training strategy introduces a shift between the RGB data and the Event data aligned with the annotations. As a result, the model prioritizes high-frequency Event data while using the rectified low-frequency RGB data as a supplement, maximizing the low-latency advantage of Event data for high-frequency detection.
More importantly, we find that these corrected Event-RGB data pairs, compared to standalone Event data, enable generalization from lower training frequency to higher inference frequencies, regardless of whether RNNs \cite{RNN_1997_TSP} or SSMs \cite{smith2022simplified} are used as memory networks. This benefit arises from the complementary information provided by the RGB data, which significantly reduces the reliance on memory networks.
This suggests that methods integrating both Event and RGB data prove more effective for high-speed object detection than those relying solely on Event data, which might intuitively seem better suited for high-speed tasks.
Additionally, in feature fusion, we introduce a shallow-feature fusion structure and a convolution-based cross-attention fusion module that achieve detection performance comparable to deep-feature fusion while considerably reducing model size, enhancing inference speed, and lowering computational costs.


\begin{figure}[!t]
	\centering
	\includegraphics[width=3.2in]{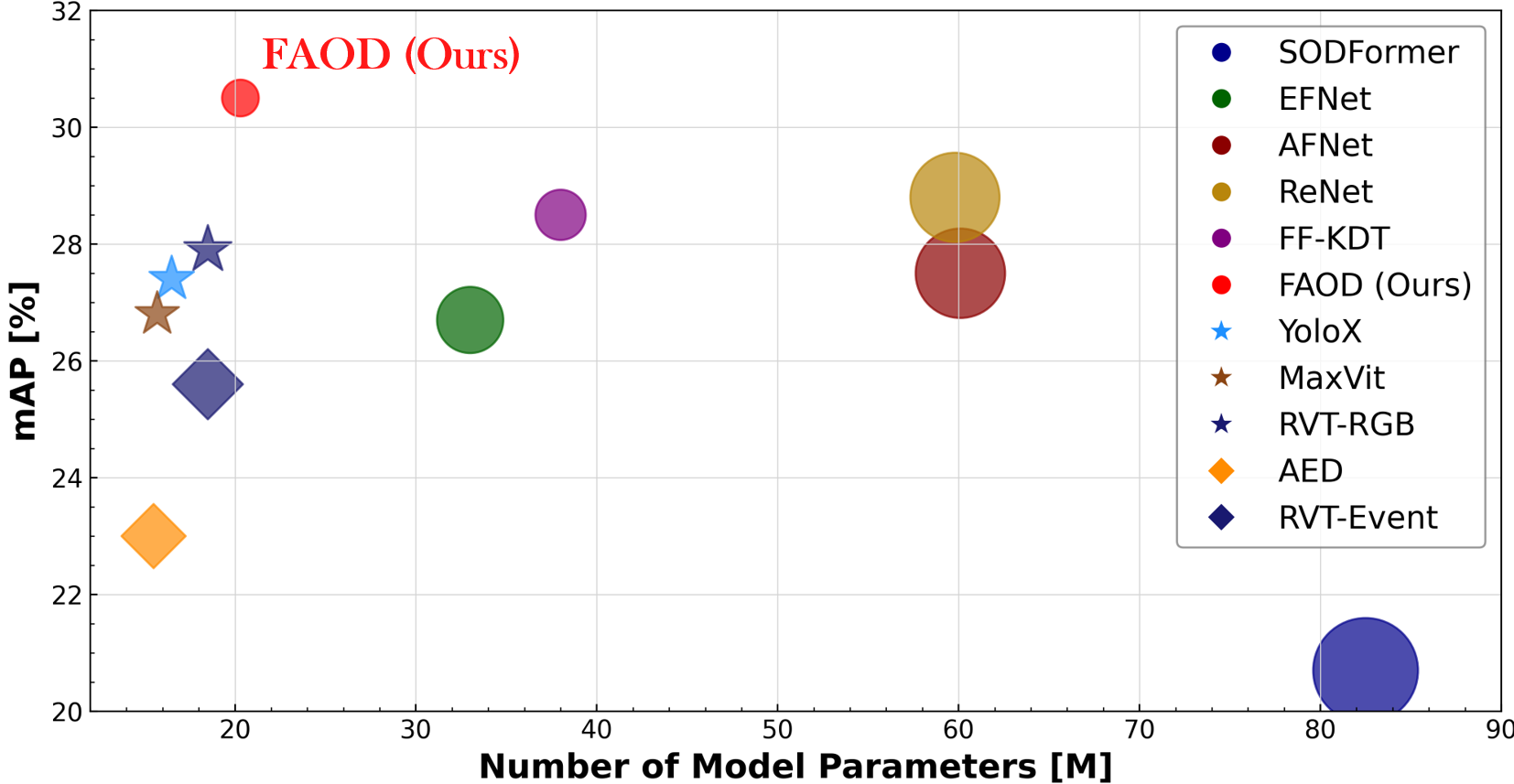}
 \vspace{-2mm}
	\caption{\textbf{Detection Performance vs. Model Parameters.} FAOD achieves the best detection performance in the PKU-DAVIS-SOD dataset with the fewest model parameters among the methods using both Events and RGB images.}
	\label{param}
 \vspace{-4mm}
\end{figure}

Our contributions can be summarized as follows:
\begin{itemize}
    \item We design a lightweight detector: FAOD for Event-RGB fusion-based object detection. FAOD achieves a remarkable trade-off between detection performance and computational cost compared to existing fusion-based detection solutions.
    \item We propose an Align Module and a training strategy Time Shift in FAOD. The Align Module performs cross-modal alignment, the Time Shift strategy not only actives the feature alignment in Align Module, but also facilitates the model's generalizability across frequency variances.
    \item Our experiments on both PKU-DAVIS-SOD and DSEC-Detection datasets demonstrate the proposed FAOD achieves the SOTA performance. Moreover, the experiments on PKU-DAVIS-SOD further demonstrate that FAOD can maintain robustness even under 80$\times$ Event-RGB frequency mismatch.
\end{itemize}

\section{Related Work}
\label{sec:formatting}

\subsection{Object Detection for Event Cameras}
Existing Event-based object detection methods \cite{event_dataset_2020_NIPS, event_spike_2022_CVPR,event_2024_SNN,event_detector_2023_TIM,event_detector_2023_CVPR, event_gnn_2021, event_gnn_2022} can be divided into two categories. 
The first directly inputs Event points, commonly employing Graph Neural Networks (GNN) \cite{event_gnn_2021, event_gnn_2022} or Spiking Neural Networks (SNN) \cite{event_spike_2020_AAAI, event_2024_SNN, event_spike_2021_TIP}. The second approach involves converting sparse Event points into dense Event frames \cite{event_dataset_2020_NIPS, event_detector_2023_CVPR, event_detector_2023_TIM, event_detector_2024_SSM} before feeding them into dense neural networks. However, those methods based on GNNs or SNNs \cite{event_2024_SNN,event_gnn_2021, event_gnn_2022, event_spike_2021_TIP} still lag behind CNN-based methods \cite{event_detector_2022_TIP, event_detector_2023_CVPR, event_detector_2024_SAST}, and some of them rely on
specific hardware support \cite{tavanaei2019deep,ghosh2009spiking,event_2024_SNN}. Therefore, compared to directly processing sparse Event points, most methods tend to convert Event points into dense frames in object detection.

To address the specific challenges posed by Event characteristics, existing methods \cite{event_dataset_2020_NIPS, event_detector_2022_TIP, event_detector_2023_CVPR, event_detector_2024_SAST, event_detector_2024_SSM} have incorporated tailored designs. 
To handle information loss caused by responding only to brightness changes of the Event camera, \cite{event_dataset_2020_NIPS, event_detector_2023_CVPR, event_detector_2022_TIP} integrate memory networks \cite{RNN_1997_TSP, LSTM_1997_NC} to provide supplementary temporal information. 
To accommodate the high temporal resolution of Event cameras, \cite{event_detector_2023_TIM, event_detector_2023_CVPR} employ simplified architectures or linear-complexity transformers for efficient processing. 
To utilize the sparsity of the Events, \cite{event_detector_2024_SAST} adopts adaptive token sampling during attention calculation to reduce computational complexity. Moreover, \cite{event_detector_2024_SSM} introduces the SSM \cite{smith2022simplified} to enhance the robustness of models under varying input frequencies. These specialized designs have significantly advanced the field of Event-based object detection.

However, due to the limited semantic information provided by Event cameras, detection performance using Event-based methods still significantly lags behind RGB-based approaches \cite{event_fusion_2023_TPAMI, li2023object, cao2024embracing}, even with the strategies above. Additionally, the finite memory length of memory networks \cite{LSTM_1997_NC, RNN_1997_TSP} introduces challenges such as memory decay during prolonged static periods and performance degradation when the network works at a frequency higher than the training frequency. These issues hinder the capabilities of Event cameras in high-speed scenarios, even when utilizing the SSM \cite{event_detector_2024_SSM,smith2022simplified}, a memory network with longer memory length. Therefore, a fusion-based approach \cite{event_fusion_2023_TPAMI,cao2024embracing} that combines Event and RGB data is the optimal choice since RGB data can compensate for the lack of semantic information in Events and alleviate the burden of memory networks to achieve frequency generalization.

\subsection{Event-RGB Fusion for Object Detection}
Most research in Event-RGB fusion for object detection aims to develop effective fusion strategies for the Event and RGB features \cite{cao2024embracing, event_fusion_2023_TPAMI, cbam, cbam_detection,effusion_2018,effusion_2019, effusion_2022}, motivated by the complementary nature of these data modalities, \textit{i.e.}, Event data excels in extreme conditions and RGB data offers rich semantic information in typical scenes. A common approach in early studies is to adopt late fusion for combining Event and RGB data. Specifically, \cite{effusion_2018} employs non-maximum suppression (NMS) to merge detection results, while \cite{effusion_2019} chooses to fuse confidence maps derived from the two modalities. More recent works \cite{event_fusion_2023_TPAMI, effusion_2022, cao2024embracing, cbam_detection} tend to explore the integration of features at the middle layers of the network, motivated by the desire to better leverage the complementary nature of different modalities while also reducing computational costs. For instance, \cite{cbam_detection} improves the convolutional attention module proposed in \cite{cbam} to enhance the shared features between Event and RGB data, while \cite{cao2024embracing, event_fusion_2023_TPAMI} utilize attention mechanisms to achieve more effective feature fusion. 

Unlike most previous studies \cite{cao2024embracing, event_fusion_2023_TPAMI, cbam_detection}, our FAOD focuses on achieving fusion between Event and RGB data across varying frequencies, tackling challenges arising from the Event-RGB Mismatch and the Train-Infer Mismatch. 
This approach, which combines Event and RGB data, allows the system to maintain high performance at very high frame rates, similar to Event-based systems, but with significantly better results than Event-based methods. Additionally, it outperforms both Event-based and RGB-based detection approaches by a considerable margin.
\section{Method}
\begin{figure*}[!t]
	\centering
	\includegraphics[width=6.7in]{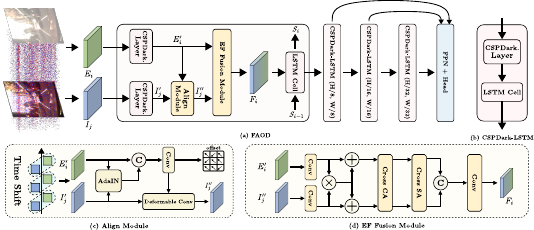}
    \vspace{-2mm}
	\caption{\textbf{Overview} of the proposed FAOD. \textbf{(a)} FAOD follows a hierarchical structure where unpaired Event frames and RGB images are fused only in the first stage. The fused features are then propagated through the subsequent three \textbf{(b)} CSPDark-LSTM Blocks for multi-scale feature extraction. The \textbf{(c)} Align Module ensures proper alignment between Event and RGB data before fusion in the \textbf{(d)} EF Fusion module. An LSTM Cell is embedded in each stage to capture temporal dependencies.}
	\label{FAOD}
 \vspace{-4mm}
\end{figure*}

\subsection{Preliminary}
\textbf{Event Representation.} Since Event cameras capture information asynchronously from individual pixels rather than capturing the entire picture at a specific timestamp, it is necessary to convert the Event stream into dense Event frames before feeding them into networks \cite{event_framing_2019_CVPR, event_framing_2019_CVPRW, event_framing_2019_ICCV, event_framing_2022_TPAMI}. 
In this work, we use a simple Event representation as used in \cite{event_detector_2023_CVPR} for all of the comparison methods to ensure a fair evaluation. 
Specifically, given a set of Event points $\Phi$ occurring within a time interval from $t_a$ to $t_b$, where each Event point $e_k$ is represented by its coordinates ($x_k$, $y_k$), timestamp $t_k$, and polarity $p_k$, \textit{i.e.}, $e_k = (x_k, y_k, p_k, t_k)$, we can obtain an Event representation $E$ using the following formula,
\begin{align}
    \label{Event representation}
    E(x, y, p, t) &= \sum_{e_k \in \Phi} \delta(p_k-p) \delta(x-x_k, y - y_k) \delta(t - \tau_k), \nonumber \\
    \tau_k &= \lfloor {\frac{t_k - t_a}{t_b - t_a} \cdot T} \rfloor,	
\end{align}
where $\delta$ indicates dirac function, $p$ represents the polarity (positive or negative), and $T$ denotes the number of discrete bins of $E$. 

\vspace{1mm}
\noindent\textbf{Spatial Feature Extraction.} We employ CSPDarkNet \cite{CSPNet}, a widely adopted backbone in the YOLO series \cite{ge2021yolox}, to extract spatial features from both Event and RGB data since it offers a compelling balance between accuracy and speed. We also do ablation studies on other spatial feature extractors such as ResNet \cite{ResNet}, MaxVit \cite{maxvit_2022_ECCV}, and Swins \cite{swins_trans} for comparison in the supplementary material.

\subsection{Main Architecture}
Fig. \ref{FAOD} illustrates the overall architecture of FAOD. 
The model processes Event frames at time step $i$ and RGB images at time step $j$, with $i \ge j$ to adapt the higher frame rate of Event data. 
In the first stage, after initial feature extraction via CSPDarkNet layers, RGB features are aligned in the Align Module using offsets generated from both Event and RGB data. 
The aligned RGB features are then fused with Event features in the EF Fusion Module, effectively combining the complementary strengths of both modalities. 
The fused features are propagated through three CSPDarkNet-LSTM Blocks to extract multi-scale features and then get the final results in the detection head. 
In our architecture, fusion is restricted to the initial stage to preserve inference speed and reduce computational complexity, while LSTM Cells are integrated into all stages to propagate temporal information effectively.

\subsection{Feature Alignment for Event-RGB Mismatch}
The higher frame rate of Event cameras compared to RGB cameras often leads to temporal and spatial misalignment when fusing, \textit{i.e.}, Event-RGB Mismatch. To maintain the high temporal resolution of Event cameras, RGB images must be warped to align with the Event frames. We use the deformable convolution \cite{dai2017deformable} to address the modality misalignment problem, as the Align Module shown in Fig. \ref{FAOD}(c). It consists of two main components: an Adaptive Instance Normalization (AdaIN) module \cite{AdaIN} for style transfer and a deformable convolution for alignment. AdaIN is employed to transfer the style of RGB images to Event frames, mitigating the impact of modality discrepancies on alignment. Specifically, AdaIN achieves style transfer by modulating the mean and variance of RGB image features as follows:
\begin{align}
    \label{AdaIN}
    \text{AdaIN}(I_{j}^{\prime}, E_{i}^{\prime}) = \sigma(E_{i}^{\prime}) \left (\frac{I_{j}^{\prime}-\mu(I_{j}^{\prime})}{\sigma(I_{j}^{\prime})} \right) + \mu(E_{i}^{\prime}),	
\end{align}
where $I_{j}^{\prime}$ and $E_{i}^{\prime}$ represent the features of the RGB image and Event frame, respectively, while $\sigma$ and $\mu$ denote the standard deviation and mean. Following AdaIN adjustment, the features extracted from the Event and RGB images are concatenated and fed into a convolution layer to produce offsets, which will guide the deformable convolution for rectifying the RGB images.

Relying solely on the Align Module is insufficient for adaptively correcting RGB images, since the Event and RGB data in the dataset are at the same fixed frequency as the annotations, leading to a lack of supervision for the deformable convolution in the Align Module. Instead of designing separate loss functions \cite{wang2024towards} or adding extra modules to guide the training of the Align Module, we propose a training strategy to achieve that. 
Specifically, we shift the RGB images in paired Event-RGB data during the input stage to simulate misalignment between the Event and RGB data. We randomly move the RGB frames forward by several time units, as Event frames typically fuse with forward RGB frames due to their higher frequency. We refer to this training strategy as \textbf{Time Shift} which can be formulated as:
\begin{equation}
\label{AdaIN}
\begin{split}
\mathcal{M}(E(t), I(t - \Delta t)) = \mathcal{M}(E(t), I(t)), \\ 
\Delta t \in [\Delta t_{\min}, \Delta t_{\max}],
\end{split}
\end{equation}
where \(\mathcal{M}(\cdot)\) represents the output of the model, and \(\Delta t\) denotes a random temporal offset within the range \([\Delta t_{\min}, \Delta t_{\max}]\), which is defined as 0 to 10 in our setting.
By generating training data with varying temporal offsets, Time Shift facilitates the training of the Align Module. Therefore, the Align Module becomes fully activated and functions effectively,  while it fails to serve its purpose solely. 
In addition, Time Shift plays a pivotal role in preventing the model from overfitting to RGB images, as detailed in Section \ref{subsec: Low-to-High Frequency Adaptation}, thereby enabling generalizability from a fixed training frequency to higher inference frequencies.

After completing the alignment, we use the cross-attention-based EF Fusion module \cite{cbam_detection} to merge the Event and RGB features, as shown in Fig. \ref{FAOD}(d). First, we multiply and add the features from both modalities to enhance their shared characteristics, establishing an initial balance. Next, by applying cross-channel and cross-spatial attention mechanisms, we facilitate deep information interactions across the channel and spatial dimensions, fully integrating the complementary features of the Events and RGB frames.

\subsection{\textbf{\normalsize Frequency Adaptation for Train-Infer Mismatch}}
\label{subsec: Low-to-High Frequency Adaptation}
Event-based detection \cite{event_dataset_2020_NIPS, event_detector_2022_TIP, event_detector_2023_CVPR, event_detector_2024_SAST, event_detector_2024_SSM} systems experience a significant performance drop when encountering data with frequencies higher than those seen during training, known as Train-Infer Mismatch, primarily due to the limited memory length and fixed-frequency adaptation of memory networks. VIT-S5 \cite{event_detector_2024_SSM} first highlighted this issue and proposed using SSMs \cite{event_detector_2024_SSM, smith2022simplified} instead of RNNs to enable generalization from low to high frequencies. 
However, we observe that within the same frequency variation from training to inference, the combination of Event information with corrected RGB data experiences significantly less performance degradation compared to standalone Event data. This suggests that FAOD facilitates generalization from lower training frequencies to higher inference frequencies.
This is because, unlike pure Event-based methods, fusion-based approaches leverage complementary RGB data, which can considerably alleviate the burden on memory networks. 

Simply incorporating RGB data into the model is insufficient. 
Additional strategies are required to prioritize Event information over RGB since generalizing from low to high frequencies primarily involves increasing the frequency of Event data, rather than RGB data. 
However, during training, the model tends to favor RGB data due to its richer semantic content, neglecting the high-frequency characteristic of the Event. 
We also address this issue through the application of Time Shift. 
In the training, the strategy introduces offsets between RGB images and annotations to encourage the network to align RGB features to Event features which synchronizes with the annotations. Consequently, during inference, the model can concentrate on utilizing both Events and aligned RGB features, rather than relying on RGB images from previous timesteps, which can maintain the low-latency nature of the Event stream toward high-frequency detection.

It is worth noting that while SSMs offer superior generalizability and longer memory compared to RNNs, their memory capacity is still limited. This limitation becomes apparent in pure Event-based detection tasks involving high-frequency Events. 
In contrast, Event-RGB fusion methods can effectively overcome this challenge by utilizing complementary information from RGB data, maintaining a better and consistent performance across varying frequencies. Therefore, we identify Event-RGB fusion as a more robust approach for high-frequency object detection.

\section{Experiments}
\begin{table*}[!t]
	\renewcommand{\arraystretch}{1.1} 
	\centering
	\caption{\textbf{Comparison results of PKU-DAVIS-SOD and DSEC-Detection datasets.} Best results in \textbf{bold}. A star $*$ indicates that this result is not directly available and estimated based on the publications. Runtime is measured in milliseconds for a batch size of 1 on an NVIDIA Tesla V100 GPU to compare against indicated timings in prior work \cite{event_fusion_2023_TPAMI}.}
	\setlength{\tabcolsep}{3.5pt} 
	\vspace{-6pt}
	\label{comparsion_exps}
	\scalebox{0.93}{
		\begin{tabular}{p{1.30cm}p{3.0cm}p{3.0cm}p{1.0cm}p{1.0cm}p{1.5cm}p{1.0cm}p{1.0cm}p{1.5cm}p{1.7cm}}
			\toprule
            & & & \multicolumn{3}{c|}{PKU-DAVIS-SOD \cite{event_fusion_2023_TPAMI}} & \multicolumn{3}{c}{DSEC-Detection \cite{gehrig2024low}} & \\
            \cline{4-9}
            
            Modality & Method & Backbone & mAP & $\text{AP}_{50}$ & Time (ms) & mAP & $\text{AP}_{50}$ & Time (ms) & Params (M) \\
            \hline
            \multirow{4}{*}{Events} & ASTMNet \cite{event_detector_2022_TIP}& (T)CNN + RNN &  -&  29.1 & 21.3 & - & - & - & $\textgreater$ 100*  \\

            & AED \cite{event_detector_2023_TIM}& CNN & 23.0 & 45.7 & 3.1 &  27.1 & 43.2 & 3.7 & 15.5 \\

            & VIT-S5 \cite{event_detector_2024_SSM}& Transformer + SSM &  23.2&  46.6& 10.7  & 23.8 & 38.7 & 11.4 & 18.2 \\
            
            & RVT \cite{event_detector_2023_CVPR}& Transformer + RNN &  25.6 & 50.3 &  7.1 & 27.7 & 44.2 & 8.0 & 18.5 \\

            \hline
            
            \multirow{5}{*}{Frame}  & YoloX \cite{ge2021yolox}& CNN & 27.4 & 50.9 & 9.3 & 38.5 & 57.8 & 9.7 & 16.5 \\

            & MaxVit \cite{maxvit_2022_ECCV}& Transformer & 26.8 & 50.5 & 5.7 & 32.8 & 51.0 & 6.5 & 15.7 \\

            & Swins \cite{swins_trans}& Transformer  & 27.7 & 52.3 & 8.4 &  34.1 & 52.0 & 9.5 & 15.8\\

            & VIT-S5 \cite{event_detector_2024_SSM}& Transformer + SSM & 28.2 & 52.2 & 10.7 & 33.2 & 49.6 & 11.6 &  18.1 \\
            
            & RVT \cite{event_detector_2023_CVPR}& Transformer + RNN   & 27.9 & 53.0 & 7.0 & 39.2 & 61.0 & 8.1 & 18.5 \\

            \hline
            \multirow{8}{*}{Fusion}  & JDF \cite{JIF} & CNN & - & 44.2 & 8.3 & - & - & - & $\textgreater$ 60* \\

            & SODFormer \cite{event_fusion_2023_TPAMI}& Transformer &  20.7 & 50.4 & 39.7 & - & - & - & 82.5\\

            & DAGr-50 \cite{gehrig2024low}& CNN + GNN &  - & - & - & 21.2 & 45.2 & 37.5 & 34.6\\

            & EFNet \cite{EFNet}& CNN + Transformer &  26.4 & 52.9 & 13.7 & 30.1 & 47.7 & 16.5 & 33.0 \\
                        
            & AFNet \cite{event_2023_tracking}& CNN &  27.5 & 53.3 & 14.5 & 31.4 & 48.8 & 19.7 & 60.1\\

            & ReNet \cite{cbam_detection}& CNN &  28.8 & 54.9 & 14.2 & 31.6 & 49.0 & 17.5 & 59.8 \\

            & FF-KDT \cite{wang2024towards}& CNN + RNN &  28.5 & 54.2 & 18.0 & 33.6 & 52.5 & 19.6 & 38.0 \\
            
            \cline{2-10}

            & FAOD (ours) & CNN + RNN  & \textbf{30.5} & \textbf{57.5} & 13.6 & \textbf{42.5} & \textbf{63.5} & 14.6 & 20.3 \\

		\toprule
	\end{tabular}}
 \vspace{-2mm}
\end{table*}

\subsection{Setup}

\textbf{Datasets } We use PKU-DAVIS-SOD \cite{event_fusion_2023_TPAMI} and DSEC-Detection \cite{gehrig2024low} as our experimental datasets. Considering the larger dataset size, manual annotations, and higher annotation frequency of PKU-DAVIS-SOD, we choose it to perform experiments addressing the Event-RGB Mismatch, Train-Inference Mismatch, and various ablation studies.

PKU-DAVIS-SOD \cite{event_fusion_2023_TPAMI}, captured by the Davis346 camera at a resolution of 346 × 260 pixels, offers perfectly aligned Event and RGB data. The dataset contains 1080.1k manually annotated bounding boxes at a rate of 25Hz, encompassing three object categories (cars, pedestrians, and two-wheelers) and Three different scene conditions (normal, motion blur, and low-light).

DSEC-Detection \cite{gehrig2024low}, acquired by the Gen3 Prophesee
The Event camera has a resolution of 640 × 480 and includes
8 categories with 39.0k annotated bounding boxes. There exist other versions of annotations for this dataset, but they only cover a subset of sequences, \textit{e.g.}, 41 sequences in \cite{effusion_2022} and 51 sequences in \cite{li2023object} with fewer bounding boxes. This work utilizes the more comprehensive annotations provided by \cite{gehrig2024low}.

\vspace{1mm}
\noindent\textbf{Implementation Details } 
We train our models using 32-bit precision for 400k iterations, employing the ADAM optimizer \cite{Adam} and a OneCycle learning rate schedule \cite{onecycle}. To enhance training efficiency, we adopt a mixed batching strategy, alternating between standard Backpropagation Through Time (BPTT) and Truncated BPTT (TBPTT) as proposed in RVT \cite{event_detector_2023_CVPR}. 
We use a batch size of 4 and a maximum learning rate of 1.5e-4. 
Both datasets are trained on a 40G A100 GPU for approximately 3 days. 
\textbf{Note that we use the complete 3 categories on the PKU-DAVIS-SOD dataset and 8 categories on the DSEC-Detection dataset during both training and testing.} 
To augment our dataset, we apply random horizontal flipping, and zooming, and use Time Shift for misaligned Event-RGB pairs testing.

We adopt mean Average Precision (mAP) \cite{COCO} as our primary evaluation metric. Additionally, to assess model efficiency, we report model parameters and inference time.

\subsection{Benchmark Comparisons}
\label{sec: Comparisons}
In this section, we comprehensively compare FAOD against SOTA methods \cite{event_detector_2023_CVPR, swins_trans, event_fusion_2023_TPAMI} on the PKU-DAVIS-SOD and DSEC-Detection datasets. Our experiments are categorized into three parts: (1) experiments on perfectly aligned Event-RGB data without Event-RGB Mismatch and Train-Infer Mismatch, (2) experiments under Event-RGB Mismatch conditions, and (3) experiments under Train-Infer Mismatch conditions. 
To ensure a thorough comparison, we include recent SOTA methods for Event-based object detection (e.g., RVT \cite{event_detector_2023_CVPR}), RGB-based tasks (e.g., Swin Transformers \cite{swins_trans}, MaxViT \cite{maxvit_2022_ECCV}), and fusion-based object detection (e.g., SODformer \cite{event_fusion_2023_TPAMI}). 
Furthermore, we introduce novel baselines by combining advanced feature extractors from other domains using Event and RGB (e.g., EFNet \cite{EFNet} for deblurring, AFNet \cite{event_2023_tracking} for tracking, FF-KDT \cite{wang2024towards} for feature point extraction) with lightweight YOLOX \cite{ge2021yolox} FPN and Head. 
Comprehensive evaluations include mAP, $\text{AP}_{50}$, inference time, and model parameters.
\begin{table}[!t]
	\renewcommand{\arraystretch}{1.1} 
	\centering
 	\caption{\textbf{Comparison results of PKU-DAVIS-SOD dataset with unpaired Event-RGB.} The frequency of Event frames is 25 Hz, whereas the RGB camera captures frames at a rate of \nicefrac{25}{N} Hz. A lower RGB frame rate results in a larger discrepancy.  “MDrop” quantifies the max performance drop of the model.}
	\setlength{\tabcolsep}{3.5pt} 
	\vspace{-8pt}
	\label{unpaired_exp}
	\scalebox{0.91}{
		\begin{tabular}{p{2.4cm}p{0.6cm}p{0.6cm}p{0.6cm}p{0.6cm}p{0.6cm}p{0.6cm}p{1.2cm}}

    \toprule 
    & \multicolumn{7}{c}{N $\colon$ RGB Frequency (\nicefrac{25}{N} Hz)} \\

    \cline{2-8}
    
    Methods  & 1 & 2 & 4 & 6 & 8 & 10 & MDrop \\

    \hline
    SODFormer \cite{event_fusion_2023_TPAMI} & 20.7 & 18.6 & 15.4 & 14.2 & 12.6 & 11.7 & 9.0 \\
    EFNet \cite{EFNet} & 26.4 & 24.8 & 21.6 & 19.8 & 16.7 & 15.9 & 10.5\\
    AFNet \cite{event_2023_tracking} & 27.5 & 25.3 & 21.2 & 20.0 & 16.5 & 15.1 & 12.4\\
    ReNet \cite{cbam_detection} & 28.8 & 27.0 & 22.8 & 21.4 & 18.3 & 17.0 & 11.8\\
    FF-KDT \cite{wang2024towards} & 28.5 & 26.7 & 22.6 & 21.1 & 17.9 & 16.5 & 12.0\\
    \hline
    FAOD (ours)& \textbf{29.7} & \textbf{29.5} & \textbf{29.2} & \textbf{29.1} & \textbf{28.6} & \textbf{28.5} & \textbf{1.2} \\
    \toprule
	
	\end{tabular}}
 \vspace{-4mm}
\end{table}
\textbf{}

\subsubsection{Experiments under Paired Event-RGB}
This experimental paradigm is commonly used in existing studies \cite{event_fusion_2023_TPAMI,cao2024embracing,cbam_detection}, serving as a benchmark for evaluating the efficacy of feature extraction and fusion strategies for these two modalities. In this setting, Event and RGB data are temporally and spatially aligned during training and testing. We conduct experiments on both the PKU-DAVIS-SOD \cite{event_fusion_2023_TPAMI} and DSEC-Detection \cite{gehrig2024low} datasets. 

\begin{table*}[!t]
	\renewcommand{\arraystretch}{1.15} 
	\centering
	\caption{\textbf{Evaluation of the model's ability to generalize from low to high frequencies.} ``Paired." refers to the mAP obtained when the model is trained and tested on perfectly aligned RGB and Event data. ``MDrop" quantifies the max performance drop of the model compared to the Paired. scenario, consisting of the drop caused by Event-RGB Mismatch and the drop caused by Train-Infer Mismatch.}
	\setlength{\tabcolsep}{3.5pt} 
	\vspace{-8pt}
	\label{low_to_high_exp}
	\scalebox{0.95}{
		\begin{tabular}{p{2.2cm}p{1.4cm}p{0.8cm}p{0.8cm}p{0.8cm}p{0.8cm}p{0.8cm}p{1.5cm}p{0.8cm}p{0.8cm}p{0.8cm}p{0.8cm}p{0.8cm}p{1.5cm}}

    \toprule
    & & \multicolumn{6}{c|}{N $\colon$ Event (25*N Hz) \& RGB (5 Hz)} 
    & \multicolumn{6}{c}{N $\colon$ Event (25*N Hz) \& RGB (2.5 Hz)}  \\

    \cline{3-14}

    Methods & Paired.& 1 & 2 & 4 & 6 & 8 & MDrop & 1 & 2 & 4 & 6 & 8 & MDrop \\
    \hline
     RVT \cite{event_detector_2023_CVPR}& 25.6 & 25.6 & 24.3 & 18.6 & 13.9 & 11.1 & 0.0+14.5 & 25.6 & 24.3 & 18.6 & 13.9 & 11.1 & 0.0+14.5 \\

     VIT-S5 \cite{event_detector_2024_SSM} & 23.2 & 23.2 & 22.4 & 20.8 & 20.0 & 19.5 & 0.0+3.7 & 23.2 & 22.4 & 20.9 & 20.0 & 19.5 & 0.0+3.7 \\

    \hline
        
    ReNet \cite{cbam_detection} & 28.8 & 21.4 & 21.4 & 21.4& 21.4& 21.4& 7.4+0.0 & 17.0 & 17.0 & 17.0 & 17.0 & 17.0 & 11.8+0.0\\

    FF-KDT \cite{wang2024towards} & 28.5 &21.1 & 20.9 & 20.6 & 20.4 & 20.3& 7.4+0.8  &16.5  & 16.5 & 16.3 & 16.2 & 16.1 & 12.0+0.4\\

    FAOD (ours) & \textbf{29.7} & \textbf{29.1} & \textbf{28.9} & \textbf{28.3} & \textbf{27.9} & \textbf{27.6} & \textbf{0.6+1.5} & \textbf{28.5} & \textbf{28.3} & \textbf{27.6} & \textbf{27.0} & \textbf{26.7} & \textbf{1.2+1.8} \\


    \toprule
	
	\end{tabular}}
 \vspace{-3mm}
\end{table*}

The experimental results in Table \ref{comparsion_exps} show that our FAOD achieves the highest mAP with few parameters while maintaining a competitive inference speed. For example, FAOD outperforms ReNet \cite{cbam_detection} by \textbf{1.7} points in terms of mAP on PKU-DAVIS-SOD \cite{event_fusion_2023_TPAMI} and achieves an impressive mAP improvement of up to \textbf{10} points on DSEC-Detection \cite{gehrig2024low}, while having only one-third of its model parameters. The results also show that FAOD achieved a \textbf{2.3} points improvement in mAP over the best-performing single-modal ViT-S5 with RGB input while incurring only an additional 2.2 M parameters and 2.9 ms inference time. This indicates that fusion-based methods have the potential to surpass single-modal methods in terms of accuracy while maintaining comparable speed and computational costs.

\subsubsection{Experiments under Event-RGB Mismatch}

In this setting, the Event data has a higher frequency than the RGB data, resulting in spatial and temporal misalignment. For instance, if the Event frames are constructed at 25 Hz and the RGB camera at 5 Hz, each RGB frame will correspond to the subsequent five timestamps of Event frames. This experiment evaluates the model's ability to handle unpaired Event-RGB data.

Table \ref{unpaired_exp} demonstrates the superiority of FAOD in handling unpaired Event-RGB data. While all the comparison methods experience a drop of approximately 10 points in performance when transitioning from fully paired to a 10-fold frequency difference, FAOD limits this degradation to \textbf{1.2} points. This robustness is attributed to the Align Module and, more importantly, the training strategy Time Shift, which provides essential training support. Although AFNet \cite{event_2023_tracking} and FF-KDT \cite{wang2024towards} also employ deformable convolutions \cite{dai2017deformable} to rectify RGB images, the lack of appropriate supervision hinders effective RGB correction.

\subsubsection{Experiments under Train-Infer Mismatch}

This setting provides a fixed frequency of RGB images (for fusion-based methods) and higher frequencies for Events than the frequency in the training. 
To maintain performance under this scenario, the model needs the ability to rectify RGB images to Events, given that the frequency of the Event frames is always higher than that of RGB images, but also requires the memory network to have generalizability from low to high frequencies. Notably, we do not use 25 Hz RGB for evaluation but use 5 Hz and 2.5 Hz instead to introduce a certain degree of offset between the RGB images and the labels, thus preventing methods from achieving high mAP by overfitting to the RGB images.

Table \ref{low_to_high_exp} presents conclusions from multiple perspectives. For Event-based methods, performance deteriorates as the frequency increases. While VIT-S5 \cite{event_detector_2024_SSM} exhibits a much smaller decline compared to RVT \cite{event_detector_2023_CVPR}, thanks to the better generalizability of the S5 block, it still experiences a significant drop of 3.7 mAP. 
ReNet \cite{cbam_detection} and FF-KDT \cite{wang2024towards}, which primarily rely on RGB information, do not suffer from deterioration as Event frequency increases, as they do not prioritize high-frequency Event data. However, they experience significant performance declines due to Event-RGB Mismatch.
FAOD, owing to its ability to correct RGB image distortions and avoid overfitting to RGB images, exhibits robust performance in various combinations of Event and RGB data at different frequencies. For instance, when supplemented with only 5 Hz RGB images, FAOD experiences a mere \textbf{1.5} points performance drop when transitioning from 25 Hz to 200 Hz Event data, less than half of the decline observed in ViT-S5. 
This highlights the robust adaptability of FAOD across different frequencies.

\subsection{Ablation Study}

This section analyzes the influence of the proposed modules and strategies on the model's performance under three scenarios: paired Event-RGB data condition, Event-RGB Mismatch, and Train-Infer Mismatch. Specifically, we investigate (1) the impact of fusion modules and memory networks on the model's performance in the paired Event-RGB scenario; (2) the effect of the Align Module and Time Shift training strategy on the model's performance in the unpaired Event-RGB scenario; and (3) the impact of different memory networks and Time Shift on the model's generalizability from low to high frequencies. We perform all the ablation studies on PKU-DAVIS-SOD \cite{event_fusion_2023_TPAMI} dataset and using the same setting in Section \ref{sec: Comparisons}.

\subsubsection{Fusion Module and Memory Network}

\textbf{EF Fusion Module} By comparing the proposed EF Fusion Module with the direct concatenation of features, we find that our module can achieve a 0.5 mAP and 1.7 $\text{AP}_{50}$ improvement with only a negligible increase of 0.1M model parameters, as shown in Table \ref{fusion_module}. These results demonstrate that the proposed module is a lightweight and effective solution for fusing Event and RGB data.
\begin{table}[!t]
	\renewcommand{\arraystretch}{1.1} 
	\centering
 	\caption{\textbf{The impacts of EF Fusion Module on the model's performance.} The EF Fusion Module leads to a higher model performance while only contributing an increase of 0.1 M parameters.}
	\setlength{\tabcolsep}{3.5pt} 
	\vspace{-8pt}
	\label{fusion_module}
	\scalebox{0.95}{
		\begin{tabular}{p{2.4 cm}p{1cm}p{1cm}p{1cm}p{2.0cm}}

    \toprule
    Fusion Methods & mAP & $\text{AP}_{50}$ &  $\text{AP}_{75}$ & Params (M)\\
    \hline

    Cat & 30.0 & 55.8 & 27.8 & \textbf{20.2}  \\

    EF Fusion & \textbf{30.5} & \textbf{57.5} & \textbf{28.5} & 20.3 \\
    \toprule
	
	\end{tabular}}
 \vspace{-3mm}
\end{table}

\begin{table}[!t]
	\renewcommand{\arraystretch}{1.0} 
	\centering
 	\caption{\textbf{The impact of LSTM Cells contributes to the overall
    model performance.} }
	\setlength{\tabcolsep}{3.5pt} 
	\vspace{-8pt}
	\label{memory_network}
	\scalebox{0.92}{
		\begin{tabular}{p{2.4 cm}p{0.8cm}p{0.8cm}p{0.8cm}p{0.8cm}p{0.8cm}p{0.8cm}}

    \toprule

     & \multicolumn{3}{c|}{FAOD} & \multicolumn{3}{c}{RVT} \\
    \cline{2-7}
    Methods & mAP & $\text{AP}_{50}$ & $\text{AP}_{75}$ & mAP & $\text{AP}_{50}$ & $\text{AP}_{75}$ \\
    \hline 
    
    w/ Reset State & 29.1 & 55.3 & 26.8 & 19.8 & 40.8 & 16.6  \\
    
    w/o Reset State & 30.5 & 57.5 & 28.5 & 25.6 & 50.3 & 22.5  \\

    \toprule
	
	\end{tabular}}
 \vspace{-7mm}
\end{table}

\vspace{1mm}
\noindent\textbf{LSTM Cell} To verify whether memory networks still provide benefits in fusion models, we evaluate the performance of FAOD and RVT \cite{event_detector_2023_CVPR} with and without the temporal information captured in the memory network. To simulate the absence of recurrent layers while maintaining the same number of parameters, we keep the model architecture identical but reset the LSTM states as used in \cite{event_detector_2023_CVPR, event_detector_2024_SSM}.

As shown in Table \ref{memory_network}, when temporal information in the memory network is removed, RVT \cite{event_detector_2023_CVPR} experiences a significant performance drop, with a decrease of 5.8 points in terms of mAP and 9.5 points in term of $\text{AP}_{50}$. In contrast, FAOD exhibits a less pronounced decline, with mAP decreasing by 1.4 points and $\text{AP}_{50}$ by 2.2 points. This suggests that the availability of RGB information can mitigate the model's reliance on the memory network. However, this does not diminish the importance of memory networks in fusion models. When RGB data becomes unreliable due to factors like overexposure or motion blur, memory networks will play a crucial role in maintaining model robustness.

\subsubsection{Ablation of RGB Alignment}
Table \ref{alignment} illustrates the impact of the Align Module and Time Shift on achieving Event-RGB alignment. The results indicate that Time Shift is crucial for enabling the model to perform RGB rectification, as it alone can sustain performance in unpaired Event-RGB scenarios. In contrast, the Align Module is ineffective when used independently due to the lack of supervision. However, when using the Align Module combined with Time Shift, the model's performance can improve further than when using Time Shift alone. 
This demonstrates that, although the model can learn to use misaligned RGB images through Time Shift without the Align Module, the incorporation of the Align Module allows for more effective utilization of RGB information, leading to improved performance.

\subsubsection{Ablation of Memory Blocks}
\begin{table}[!t]
	\renewcommand{\arraystretch}{1.1} 
	\centering
 	\caption{\textbf{The impacts of Align Module (AM) and Time Shift to the model's ability of RGB images alignment.} Both parts are important for the model to align effectively.}
	\setlength{\tabcolsep}{3.5pt} 
	\vspace{-8pt}
	\label{alignment}
	\scalebox{0.93}{
		\begin{tabular}{p{0.6 cm}p{1.5cm}p{.6cm}p{0.6cm}p{0.6cm}p{0.6cm}p{0.6cm}p{0.6cm}p{1.0cm}}

    \toprule
    & & \multicolumn{7}{c}{N $\colon$ RGB Frequency (\nicefrac{25}{N} Hz)} \\

    \cline{3-9}
    
    AM & Time Shift & 1 & 2 & 4 & 6 & 8 & 10 & Drop \\
    \hline
    &  & 30.3 & 28.1  & 23.3 & 21.6 & 17.9 & 16.3 & 14.0 \\
    \checkmark & & 30.5 & 28.4 & 23.6 & 21.8 & 18.0 & 16.6 & 13.9 \\
    & \checkmark  & 28.5 & 28.1 & 27.8 & 27.5 & 27.3 & 27.1 & 1.4  \\
    \checkmark & \checkmark  & \textbf{29.7} & \textbf{29.5} & \textbf{29.2} & \textbf{29.1} & \textbf{28.6} & \textbf{28.5} & \textbf{1.2}  \\

    \toprule
	
	\end{tabular}}
 \vspace{-2mm}
\end{table}

\begin{table}[!t]
	\renewcommand{\arraystretch}{1.1} 
	\centering
 	\caption{\textbf{The impacts of memory networks and Time Shift to the model's ability of frequency generalization for low to high.} Time shift enables the model to achieve frequency generalization, whether using RNN or SSM.}
	\setlength{\tabcolsep}{3.5pt} 
	\vspace{-8pt}
	\label{frequency}
	\scalebox{0.90}{
		\begin{tabular}{p{1.0 cm}p{0.5cm}p{1.5cm}p{0.6cm}p{0.6cm}p{0.6cm}p{0.6cm}p{0.6cm}p{1.0cm}}

    \toprule
    & & & \multicolumn{6}{c}{N $\colon$ Event (25*N Hz), RGB (2.5 Hz)} \\

    \cline{4-9}
    
    LSTM & S5 & Time Shift & 1 & 2 & 4 & 6 & 8 & MDrop \\
    \hline
    \checkmark &  &  & 16.6 & 16.6 & 16.5 & 16.4 & 16.3 & 14.2  \\
     & \checkmark &  & 15.7 & 15.7 & 15.7 & 15.7 & 15.6 & 14.5 \\
    \checkmark &  & \checkmark & \textbf{28.5} & \textbf{28.3} & \textbf{27.6} & \textbf{27.0} & \textbf{26.7} & 3.0  \\
     & \checkmark & \checkmark & 27.6 & 27.3 & 26.9 & 26.6 & 26.5 & \textbf{2.4}  \\

    \toprule
	
	\end{tabular}}
 \vspace{-2mm}
\end{table}

This ablation investigates the impact of different memory networks and Time Shift on the model's ability to generalize from low to high frequencies. As shown in Table \ref{frequency}, both LSTM and S5 blocks can achieve generalization when supplemented with RGB and Time Shift training strategy. Without Time Shift, the model will overfit RGB data, significantly dropping overall mAP. Additionally, we observe that while SSM exhibited slightly better generalization capabilities than LSTM, it lags behind LSTM regarding overall mAP, inference speed, and training stability. 

\section{Conclusion}

To enable high-speed object detection by integrating both Events and RGB frames, we propose a novel fusion detection model named FAOD, to address two critical mismatches, \textit{i.e.}, Event-RGB Mismatch and Train-Infer Mismatch. 
By employing the Align Module and Time Shift training strategy, FAOD enables RGB rectification even when only paired Event-RGB data is available for training. 
Moreover, we find that the data composed of high-frequency Events and wrapped low-frequency RGB images used in FAOD, demonstrates improved generalizability from low to high frequencies compared to the standalone Event data. This suggests that the approaches based on fusion are better suited for high-speed detection scenarios than those relying solely on Events. 
In addition, thanks to the shallow feature fusion structure and effective modules, FAOD achieves low-latency inference while maintaining high detection performance. 
Experimental results demonstrate that FAOD can not only achieve SOTA performance at fixed training frequency with effectively balanced accuracy, speed, and parameter efficiency, but also can be utilized to fuse Event and RGB data at arbitrary frequencies.
\clearpage
\maketitlesupplementary
\definecolor{DeepBlue}{rgb}{0.1, 0.2, 0.6}
\newcommand{\question}[1]{\textbf{\textcolor{Black}{#1}}}
\newcommand{\answer}[1]{\textit{\textcolor{DeepBlue}{A: #1}}}
\setcounter{page}{1}

\noindent \textbf{The supplementary materials contain the following:}
\begin{itemize}
    \item[-] Shallow Fusion vs. Deep Fusion (\textbf{Section \ref{sec: shallow_deep}}).
    \item[-] The Supervision of Align Module (\textbf{Section \ref{sec: Aligned_loss}}).
    \item[-] Ablation of Spatial Feature Extractor (\textbf{Section \ref{sec: Extractor}}).
    \item[-] Visualization Results (\textbf{Section \ref{sec: Visualization}}).
    \item[-] Question \& Answers (\textbf{Section \ref{sec:Question}}).
\end{itemize}

\section{Shallow Fusion vs. Deep Fusion}
\label{sec: shallow_deep}
\begin{figure}[h]
	\centering
	\includegraphics[width=3.3in]{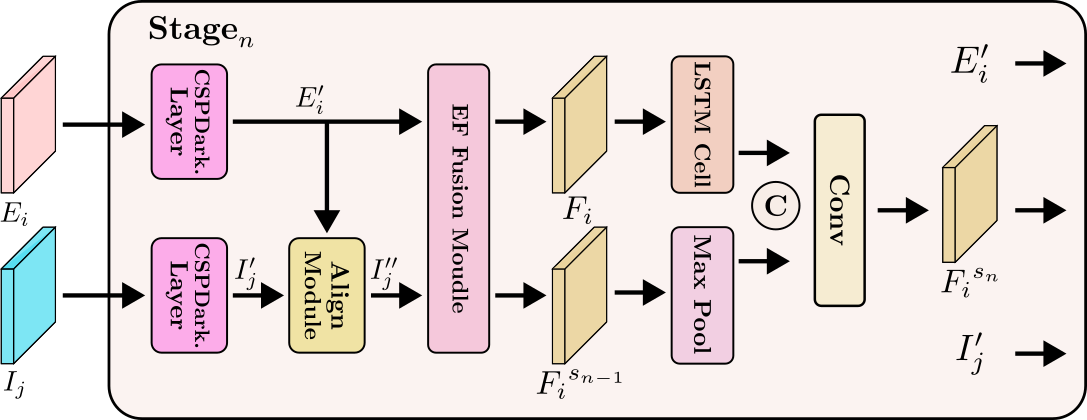}
	\caption{\textbf{Deep feature fusion framework.} Our deep-level fusion network comprises four stages, each incorporating Event and RGB modalities and the fused features from the preceding stage. The final fused features are fed into an FPN and detection head for object detection.}
	\label{deep}
\end{figure}

This section analyzes the advantages of shallow feature fusion over deep feature fusion architecture. 
We extend the shallow feature fusion architecture that applies to the first stage to a deep feature fusion architecture that applies to all stages, as shown in Figure \ref{deep}. 
Unlike the shallow feature fusion architecture, the deep feature fusion architecture feeds both the Event and RGB modalities, as well as the fused features, into all three later stages. 
Specifically, in the deep feature fusion architecture, the fused features are concatenated with the features from the previous stage. To ensure consistent scales, the features of the prior stage undergo a max-pooling operation. Moreover, in the deep feature fusion architecture, the feature extractors for both the Event and RGB modalities are completely independent, with each stage including an Align Module and an EF Fusion module, theoretically enabling more comprehensive utilization of both types of features.

\begin{table}[h]
	\renewcommand{\arraystretch}{1.2} 
	\centering
 	\caption{\textbf{Shallow Fusion vs. Deep Fusion.} The framework based on shallow feature fusion achieves a better trade-off between accuracy and model complexity.}
	\setlength{\tabcolsep}{3.5pt} 
	\vspace{-8pt}
	\label{fusion_type}
	\scalebox{0.95}{
		\begin{tabular}{p{1.4 cm}p{1.2cm}p{1.2cm}p{1.2cm}p{1.2cm}p{1.2cm}}

    \toprule
    Type & mAP & $\text{AP}_{50}$ &  $\text{AP}_{75}$ & Params & Time \\
    \hline

    Deep. & 28.7 & 54.0 & 26.8 & 55.4 & 26.9 \\

    Shallow. & \textbf{30.5} & \textbf{57.5} & \textbf{28.5} & \textbf{20.3} &\textbf{13.6} \\
    \toprule
	
	\end{tabular}}
\end{table}

As shown in Table \ref{fusion_type}, we compare the two fusion types in terms of mAP, $\text{AP}_{50}$, $\text{AP}_{75}$, model parameters, and inference time on a single V100 GPU. The results indicate the deep feature fusion architecture exhibits a noticeable decline in mAP, $\text{AP}_{50}$, and $\text{AP}_{75}$, with drops of 1.3, 3.0, and 1.2 points, respectively. These results indicate that deep feature fusion does not offer advantages for accuracy.  Furthermore, the deep feature fusion architecture has a significantly larger model size of 55.4M parameters, more than double that of the shallow feature fusion architecture, and its inference speed is halved. Considering these observations, we ultimately opt for the shallow feature fusion architecture.

\section{The Supervision of Align Module}
\label{sec: Aligned_loss}
\begin{figure}[h]
	\centering
	\includegraphics[width=3.2in]{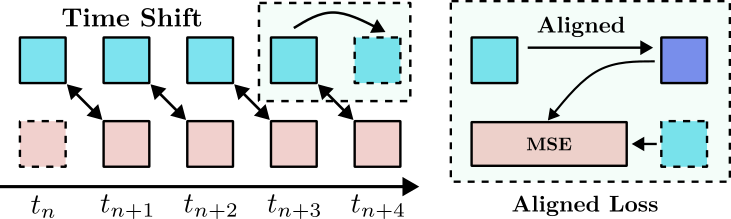}
	\caption{\textbf{Caculation of Aligned Loss.} Aligned Loss is computed based on RGB frames that have passed through the Align Module under unpaired conditions and the RGB frames perfectly aligned with the Event frames. This loss serves as supervision for the Align Module.}
	\label{supervision}
\end{figure}

This section investigates whether providing direct supervision to the Align Module is necessary. Under the Time Shift training strategy condition, the Event and RGB data are misaligned. However, since we have access to the entire sequence during training, we can obtain RGB data that is perfectly aligned with the Event data. Consequently, we can directly supervise the Align Module by calculating the Mean Squared Error (MSE) loss between the corrected RGB data and the perfectly aligned RGB data, as illustrated in Figure \ref{supervision}. The loss function of the Align Module is:

\begin{align}
    \label{Align_loss_fun}
    \mathcal{L}_{AM} = MSE(\text{AM}(F_{unpaired}), F_{paired}),
\end{align}
where $AM$ indicates the Align Module, $F_{unpaired}$ refers to the RGB feature with a temporal offset relative to the Event feature, while $F_{paired}$ denotes the RGB feature that is temporally aligned with the Event feature.

\begin{figure*}[h]
	\centering
	\includegraphics[width=7.0in]{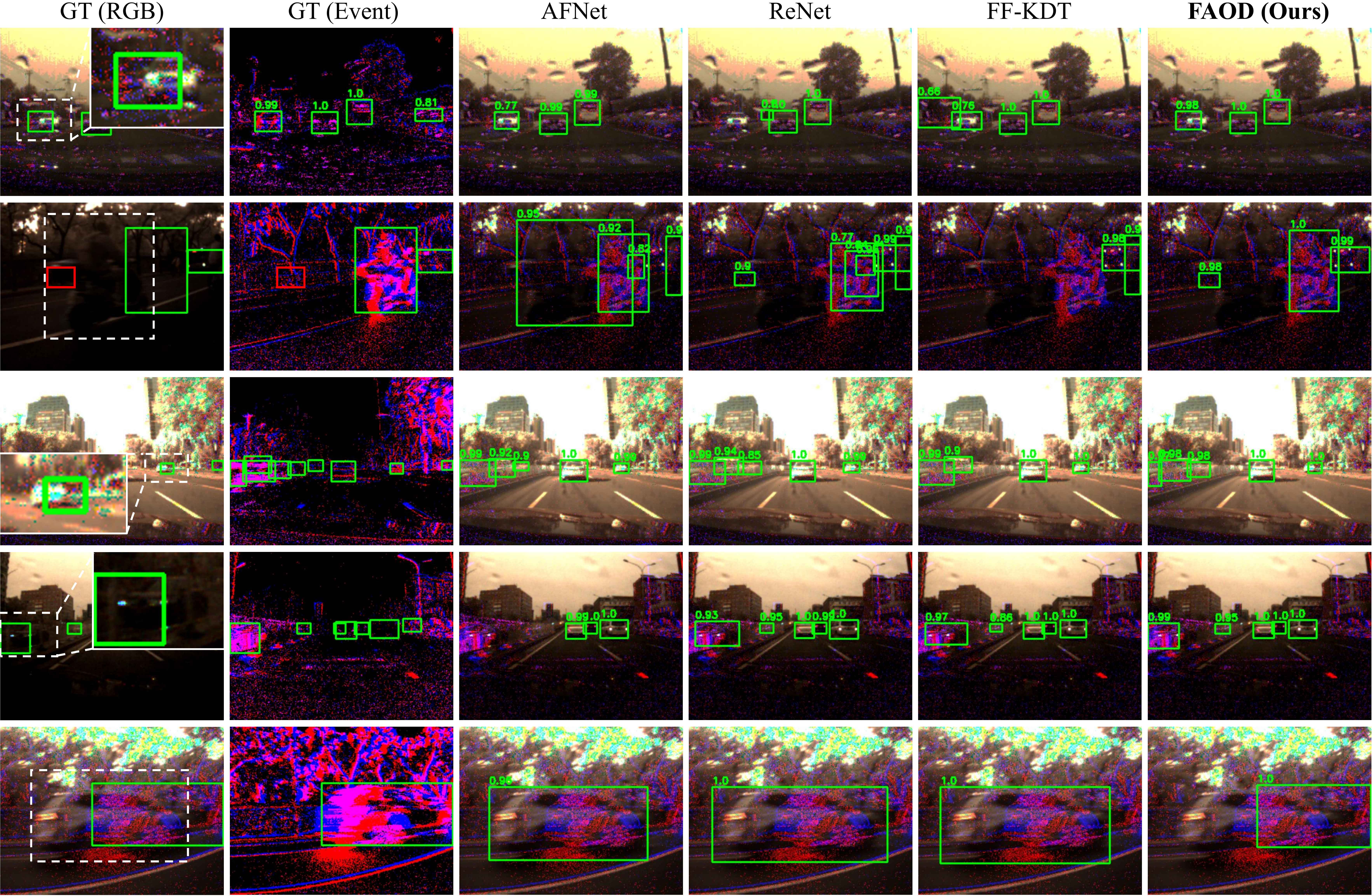}
	\caption{\textbf{Visualizations of AFNet, ReNet, FF-KDT, and FAOD on unpaired Event-RGB data.} The Events are captured at 25 Hz, while the RGB images are captured at 2.5 Hz, leading to temporal and spatial bias. Therefore, the obtained results should align with the Events, while exhibiting some offset relative to the RGB images. The red bounding boxes represent additional annotations missed in the original GT. }
	\label{compare_unpair}
    \vspace{-4mm}
\end{figure*}

\begin{figure*}[h]
	\centering
	\includegraphics[width=7.0in]{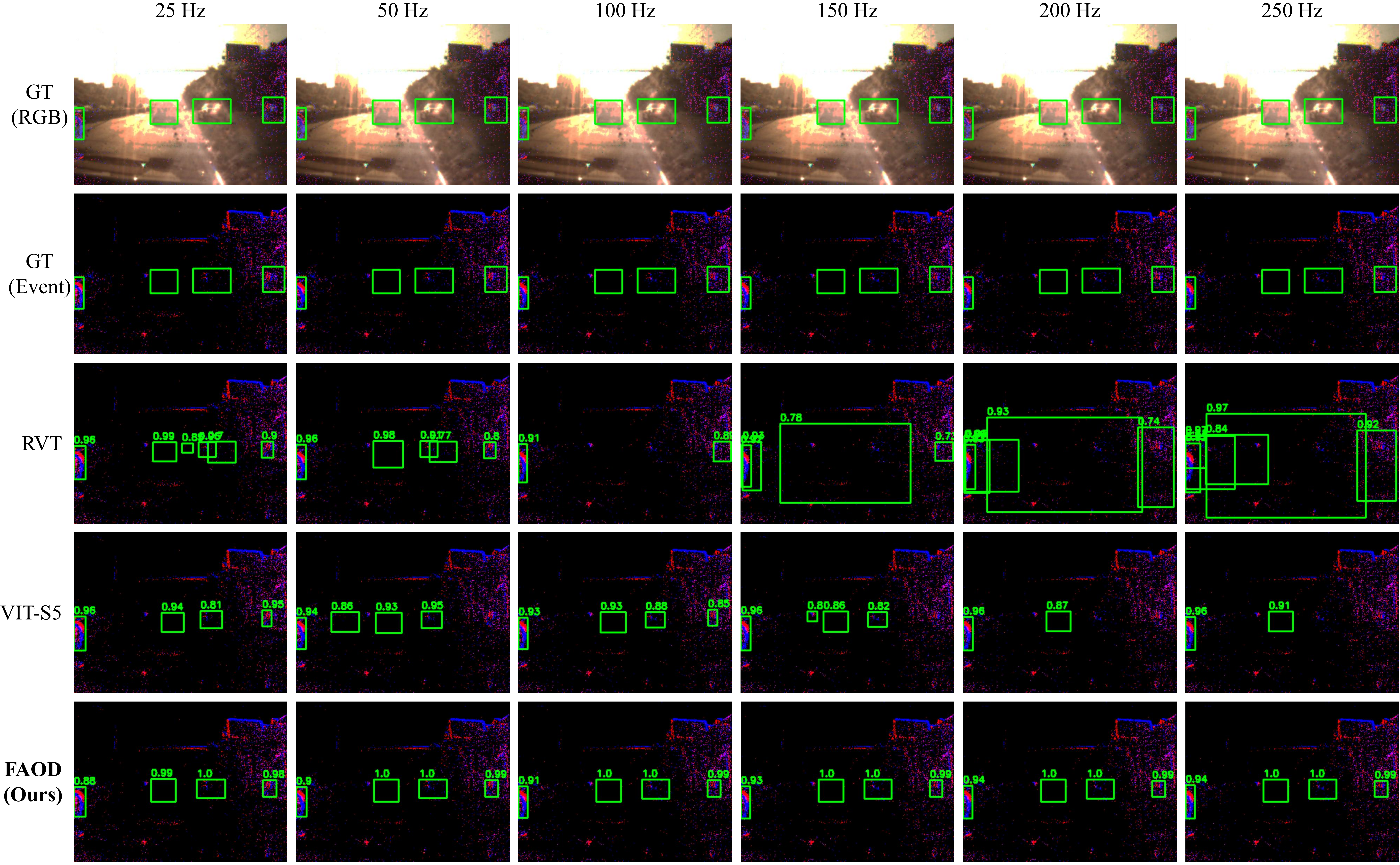}
	\caption{\textbf{Visualizations of performance for RVT, ViT-S5, and FAOD under varying Event data frequencies.} The RGB frequency used by FAOD is 2.5 Hz.}
	\label{compare_freq}
\end{figure*}

\begin{table}[h]
	\renewcommand{\arraystretch}{1.2} 
	\centering
 	\caption{\textbf{The impact of Align loss on model performance.}  The explicit supervision provided by Aligned loss does not yield performance gains.}
	\setlength{\tabcolsep}{3.5pt} 
	\vspace{-8pt}
	\label{unpaired_loss_exp}
	\scalebox{0.91}{
		\begin{tabular}{p{2cm}p{0.6cm}p{0.6cm}p{0.6cm}p{0.6cm}p{0.6cm}p{0.6cm}p{1.2cm}}

    \toprule 
    & \multicolumn{7}{c}{N $\colon$ RGB Frequency (\nicefrac{25}{N} Hz)} \\

    \cline{2-8}
    
    Align Loss  & 1 & 2 & 4 & 6 & 8 & 10 & MDrop \\

    \hline
    w  & 29.4 & 29.3 & 29.0 & 28.9 & \textbf{28.7} & 28.5 & \textbf{0.9} \\
    w/o  & \textbf{29.7} & \textbf{29.5} & \textbf{29.2} & \textbf{29.1} & 28.6 & 28.5 & 1.2\\

    \toprule
	
	\end{tabular}}
\end{table}

As shown in Table \ref{unpaired_loss_exp}, the experimental results indicate that adding Aligned Loss does not improve model performance but leads to a slight degradation. This suggests that forcibly correcting misaligned images to match images aligned perfectly may have adverse effects, and using implicit supervision to allow the model to learn how to utilize RGB information automatically is a better approach.

\section{Ablation of Spatial Feature Extractor}
\label{sec: Extractor}
In this chapter, we evaluate the impact of various feature extractors on model performance. Specifically, we compare CNN-based architectures such as CSPDarknet \cite{CSPNet} and ResNet \cite{ResNet}, as well as Transformer-based models like MaxViT \cite{maxvit_2022_ECCV} and Swins Transformer \cite{swins_trans}. To ensure a fair comparison, we control for model size by keeping the number of parameters approximately equal across different extractors. Our evaluation uses multiple metrics, including mAP, $\text{AP}_{50}$, $\text{AP}_{75}$, inference speed, and model size. The results are shown in Table \ref{spatial_extractor}.

\begin{table}[h]
	\renewcommand{\arraystretch}{1.2} 
	\centering
 	\caption{\textbf{The impacts of different spatial feature extractors on the model's performance.} We evaluate CNN-based architectures such as CSPDarkNet and ResNet, along with Transformer-based models like MaxViT and Swin Transformer.}
	\setlength{\tabcolsep}{3.5pt} 
	\vspace{-8pt}
	\label{spatial_extractor}
	\scalebox{0.95}{
		\begin{tabular}{p{2.6 cm}p{0.9cm}p{0.9cm}p{0.9cm}p{1cm}p{0.9cm}}

    \toprule
    Extractor-type & mAP & $\text{AP}_{50}$ & $\text{AP}_{75}$ & Params& Time \\
    \hline

    ResNet \cite{ResNet} & 28.4 & 52.8 & 26.7 & 22.2 & \textbf{6.2} \\

    MaxVit \cite{maxvit_2022_ECCV} & 29.4 & 55.2 & 27.8 & 20.3 & 10.6\\

    Swins \cite{swins_trans} & 30.2 & 56.5 & 28.2 & \textbf{19.6} & 17.2\\

    CSPDarkNet \cite{CSPNet} & \textbf{30.5} & \textbf{57.5} & \textbf{28.5} & 20.3  & 13.6 \\

    \toprule
	
	\end{tabular}}
\end{table}

Experimental results demonstrate that CSPDarknet achieves the highest performance on metrics such as mAP, $\text{AP}_{50}$, and $\text{AP}_{75}$ among all feature extractors. Swins Transformer exhibits comparable performance but suffers from a longer inference time. ResNet and MaxViT offer relatively faster inference speeds but fall short of CSPDarknet in terms of mAP. Considering both performance and speed, we ultimately select CSPDarknet as our feature extractor.

\section{Visualization Results}
\label{sec: Visualization}
This section presents the visual results, including the performance of different methods under Event-RGB Mismatch (as shown in Figure \ref{compare_unpair}) and the generalizability of pure Event methods and fusion-based approaches under Train-Infer Mismatch (as shown in Figure \ref{compare_freq}).

\vspace{1mm}
\noindent\textbf{Visualization under Event-RGB Mismatch.} Figure \ref{compare_freq} visualizes the localization results, revealing that AFNet \cite{event_2023_tracking}, ReNet \cite{cbam_detection}, and FF-KDT \cite{wang2024towards} predominantly place bounding boxes on lagged RGB frames rather than on the higher-frequency Event data. In contrast, FAOD is primarily driven by Event data, with bounding boxes aligning well with the corresponding Events. Furthermore, we observe that these fusion-based methods do not strictly adhere to RGB data but tend to select the higher-quality modality. While this mechanism is effective in many scenarios, such as fusing RGB and infrared images \cite{RGB_infrared} for day-night complementary, it is unsuitable for Event-RGB data with disparate frequencies. In such cases, focusing primarily on Event data is essential for achieving high-frequency object detection.

\vspace{1mm}
\noindent\textbf{Visualization under Train-Infer Mismatch.} Figure \ref{compare_freq} presents the visualization of RVT \cite{event_detector_2023_CVPR}, ViT-S5 \cite{event_detector_2024_SSM}, and FAOD, all equipped with memory networks, on Event data with varying frequencies. It is evident that both RVT and ViT-S5 experience increased false positives and missed detections as the Event frequency rises, with ViT-S5 showing relatively better performance. In contrast, FAOD maintains a stable output across different frequencies. This highlights the superior generalizability of FAOD from low to high frequencies compared to Event-only methods, suggesting its suitability for high-frequency object detection.

More video visualization can be found in \url{https://anonymous.4open.science/r/FAOD-master}.

\section{Questions \& Answers}
\label{sec:Question}

\begin{enumerate}[label=\textbf{\textcolor{black}{Q\arabic*:}}, wide, labelwidth=!, labelindent=0pt]
    \item \question{Why using a series of misaligned datasets created by Time shift for training allows the model to handle Event-RGB Mismatch?}
    
    \noindent \answer{The challenge of the Event-RGB Mismatch lies in the temporal and spatial misalignment between Event and RGB frames. Event and RGB data are typically perfectly aligned in the previous training paradigm. As a result, the model does activate the Align Module to correct RGB data during training, which subsequently renders the Align Module ineffective at rectifying misaligned RGB data during inference.\\
    The essence of the proposed Time Shift strategy is to simulate data under conditions resembling the Event-RGB Mismatch during inference. While Time Shift generates scenarios where RGB data lags behind Event data, and Event-RGB Mismatch involves differing sampling frequencies, the core issue in both cases is identical: temporal and spatial misalignment between Event and RGB frames.}

    \vspace{2mm}
    \item \question{It can be found from the experimental results (Table 3 in the main text) that ReNet and FF-KDT do not suffer serious performance degradation due to Train-Infer Mismatch, does that mean all fusion-based models can overcome the Train-Infer Mismatch?}
    
    \noindent \answer{For fusion-based methods, addressing the Event-RGB Mismatch is a prerequisite for evaluating their ability to resolve the Train-Infer Mismatch. In previous approaches using Events and RGB images, RGB is considered the primary modality, which inherently prevents these methods from achieving high-frequency detection. Even when the frequency of Event data increases from 25 Hz to 200 Hz, the Event data only plays the role of auxiliary data anchored on the RGB data.
    As a result, they fail to perform inference at frequencies higher than the training frequency, rendering them incapable of addressing the Train-Infer Mismatch.}

    \vspace{2mm}
    \item \question{How does Time Shift affect model performance and its usage in different scenarios?}
    
    \noindent \answer{In the experiments on paired Event-RGB data, omitting the Time Shift results in identical data forms during training and inference, thereby achieving a slightly higher mAP score. However, introducing Time Shift leads to only a minor decrease of 0.8 points, which is an acceptable trade-off. On the other hand, in most scenarios, the need for Event and RGB at different frequencies during inference is typically predictable. Therefore, the decision to employ Time Shift can be flexibly made based on specific application requirements.}

    \vspace{2mm}
    \item \question{Why is the fusion method more suitable for high-frequency object detection than the pure Event method?}
    
    \noindent  \answer{On one hand, by leveraging lightweight architectures and modules, fusion-based models can achieve computational efficiency and inference speeds comparable to single-modal models. On the other hand, when generalizing to higher-frequency Event data, fusion-based approaches demonstrate significantly greater robustness and superior detection performance compared to Event-only models. Therefore, we believe that fusion-based solutions are better suited for high-speed, high-frequency detection scenarios.}
\end{enumerate}

{
    \small
    \bibliographystyle{ieeenat_fullname}
    \bibliography{main}

\begin{thebibliography}{47}
\providecommand{\natexlab}[1]{#1}
\providecommand{\url}[1]{\texttt{#1}}
\expandafter\ifx\csname urlstyle\endcsname\relax
  \providecommand{\doi}[1]{doi: #1}\else
  \providecommand{\doi}{doi: \begingroup \urlstyle{rm}\Url}\fi

\bibitem[Baldwin et~al.(2022)Baldwin, Liu, Almatrafi, et~al.]{event_framing_2022_TPAMI}
R~Wes Baldwin, Ruixu Liu, Mohammed Almatrafi, et~al.
\newblock Time-ordered recent event (tore) volumes for event cameras.
\newblock \emph{{IEEE Trans. Pattern Anal. Mach. Intell.}}, 45\penalty0 (2):\penalty0 2519--2532, 2022.

\bibitem[Cannici et~al.(2019)Cannici, Ciccone, Romanoni, and Matteucci]{event_framing_2019_CVPRW}
Marco Cannici, Marco Ciccone, Andrea Romanoni, and Matteo Matteucci.
\newblock Asynchronous convolutional networks for object detection in neuromorphic cameras.
\newblock In \emph{{Proc. IEEE/CVF Conf. Comput. Vis. Pattern Recognit. workshops}}, pages 0--0, 2019.

\bibitem[Cao et~al.(2024)Cao, Zhang, Xia, et~al.]{cao2024embracing}
Hu Cao, Zehua Zhang, Yan Xia, et~al.
\newblock Embracing events and frames with hierarchical feature refinement network for object detection.
\newblock \emph{arXiv preprint arXiv:2407.12582}, 2024.

\bibitem[Chakraborty et~al.(2021)Chakraborty, She, and Mukhopadhyay]{event_spike_2021_TIP}
Biswadeep Chakraborty, Xueyuan She, and Saibal Mukhopadhyay.
\newblock A fully spiking hybrid neural network for energy-efficient object detection.
\newblock \emph{{IEEE Trans. Image Process}}, 30:\penalty0 9014--9029, 2021.

\bibitem[Chen(2018)]{effusion_2018}
Nicholas~FY Chen.
\newblock Pseudo-labels for supervised learning on dynamic vision sensor data, applied to object detection under ego-motion.
\newblock In \emph{{Proc. IEEE/CVF Conf. Comput. Vis. Pattern Recognit. workshops}}, pages 644--653, 2018.

\bibitem[Dai et~al.(2017)Dai, Qi, Xiong, et~al.]{dai2017deformable}
Jifeng Dai, Haozhi Qi, Yuwen Xiong, et~al.
\newblock Deformable convolutional networks.
\newblock In \emph{{Proc IEEE Int. Conf. on Computer Vision}}, pages 764--773, 2017.

\bibitem[Diederik(2014)]{Adam}
P~Kingma Diederik.
\newblock Adam: A method for stochastic optimization.
\newblock \emph{{Proc Int. Conf. on Learn. Repre.}}, 2014.

\bibitem[Fan et~al.(2024)Fan, Zhang, Liu, Li, and Lu]{event_2024_SNN}
Yimeng Fan, Wei Zhang, Changsong Liu, Mingyang Li, and Wenrui Lu.
\newblock Sfod: Spiking fusion object detector.
\newblock In \emph{{Proc. IEEE/CVF Conf. Comput. Vis. Pattern Recognit.}}, pages 17191--17200, 2024.

\bibitem[Gallego et~al.(2020)Gallego, Delbr{\"u}ck, Orchard, et~al.]{event_survey_2020_TPAMI}
Guillermo Gallego, Tobi Delbr{\"u}ck, Garrick Orchard, et~al.
\newblock Event-based vision: A survey.
\newblock \emph{{IEEE Trans. Pattern Anal. Mach. Intell.}}, 44\penalty0 (1):\penalty0 154--180, 2020.

\bibitem[Ge et~al.(2021)Ge, Liu, Wang, et~al.]{ge2021yolox}
Zheng Ge, Songtao Liu, Feng Wang, et~al.
\newblock Yolox: Exceeding yolo series in 2021.
\newblock \emph{arXiv preprint arXiv:2107.08430}, 2021.

\bibitem[Gehrig and Scaramuzza(2024)]{gehrig2024low}
Daniel Gehrig and Davide Scaramuzza.
\newblock Low-latency automotive vision with event cameras.
\newblock \emph{Nature}, 629\penalty0 (8014):\penalty0 1034--1040, 2024.

\bibitem[Gehrig and Scaramuzza(2023)]{event_detector_2023_CVPR}
Mathias Gehrig and Davide Scaramuzza.
\newblock Recurrent vision transformers for object detection with event cameras.
\newblock In \emph{{Proc. IEEE/CVF Conf. Comput. Vis. Pattern Recognit.}}, pages 13884--13893, 2023.

\bibitem[Ghosh-Dastidar and Adeli(2009)]{ghosh2009spiking}
Samanwoy Ghosh-Dastidar and Hojjat Adeli.
\newblock Spiking neural networks.
\newblock \emph{International journal of neural systems}, 19\penalty0 (04):\penalty0 295--308, 2009.

\bibitem[He et~al.(2016)He, Zhang, Ren, and Sun]{ResNet}
Kaiming He, Xiangyu Zhang, Shaoqing Ren, and Jian Sun.
\newblock Deep residual learning for image recognition.
\newblock In \emph{{Proc. IEEE/CVF Conf. Comput. Vis. Pattern Recognit.}}, pages 770--778, 2016.

\bibitem[Hochreiter and Schmidhuber(1997)]{LSTM_1997_NC}
Sepp Hochreiter and Jürgen Schmidhuber.
\newblock Long short-term memory.
\newblock \emph{Neural Computation}, 9\penalty0 (8):\penalty0 1735--1780, 1997.

\bibitem[Huang and Belongie(2017)]{AdaIN}
Xun Huang and Serge Belongie.
\newblock Arbitrary style transfer in real-time with adaptive instance normalization.
\newblock In \emph{{Proc IEEE Int. Conf. on Computer Vision}}, pages 1501--1510, 2017.

\bibitem[Jiang et~al.(2019)Jiang, Xia, Huang, et~al.]{effusion_2019}
Zhuangyi Jiang, Pengfei Xia, Kai Huang, et~al.
\newblock Mixed frame-/event-driven fast pedestrian detection.
\newblock In \emph{{Proc IEEE Int. Conf. on Robot. and Automat.}}, pages 8332--8338. IEEE, 2019.

\bibitem[Kim et~al.(2020)Kim, Park, Na, and Yoon]{event_spike_2020_AAAI}
Seijoon Kim, Seongsik Park, Byunggook Na, and Sungroh Yoon.
\newblock Spiking-yolo: spiking neural network for energy-efficient object detection.
\newblock In \emph{{Proc AAAI Conf. on Arti. Intell.}}, pages 11270--11277, 2020.

\bibitem[Li et~al.(2023{\natexlab{a}})Li, Li, and Tian]{event_fusion_2023_TPAMI}
Dianze Li, Jianing Li, and Yonghong Tian.
\newblock Sodformer: Streaming object detection with transformer using events and frames.
\newblock \emph{{IEEE Trans. Pattern Anal. Mach. Intell.}}, 45\penalty0 (11):\penalty0 14020--14037, 2023{\natexlab{a}}.

\bibitem[Li et~al.(2019)Li, Dong, Yu, et~al.]{JIF}
Jianing Li, Siwei Dong, Zhaofei Yu, et~al.
\newblock Event-based vision enhanced: A joint detection framework in autonomous driving.
\newblock In \emph{2019 IEEE International Conference on Multimedia and Expo}, pages 1396--1401, 2019.

\bibitem[Li et~al.(2022)Li, Li, Zhu, et~al.]{event_detector_2022_TIP}
Jianing Li, Jia Li, Lin Zhu, et~al.
\newblock Asynchronous spatio-temporal memory network for continuous event-based object detection.
\newblock \emph{{IEEE Trans. Image Process}}, 31:\penalty0 2975--2987, 2022.

\bibitem[Li et~al.(2023{\natexlab{b}})Li, Liniger, Millhaeusler, et~al.]{li2023object}
Lei Li, Alexander Liniger, Mario Millhaeusler, et~al.
\newblock Object-centric cross-modal feature distillation for event-based object detection.
\newblock \emph{arXiv preprint arXiv:2311.05494}, 2023{\natexlab{b}}.

\bibitem[Li et~al.(2021)Li, Niklaus, Snavely, and Wang]{event_gnn_2021}
Zhengqi Li, Simon Niklaus, Noah Snavely, and Oliver Wang.
\newblock Neural scene flow fields for space-time view synthesis of dynamic scenes.
\newblock In \emph{{Proc. IEEE/CVF Conf. Comput. Vis. Pattern Recognit.}}, pages 6498--6508, 2021.

\bibitem[Lin et~al.(2014)Lin, Maire, Belongie, et~al.]{COCO}
Tsung-Yi Lin, Michael Maire, Serge Belongie, et~al.
\newblock Microsoft coco: Common objects in context.
\newblock In \emph{{Proc Eur. Conf. on Computer Vision}}, pages 740--755. Springer, 2014.

\bibitem[Liu et~al.(2023)Liu, Xu, Yang, et~al.]{event_detector_2023_TIM}
Bingde Liu, Chang Xu, Wen Yang, et~al.
\newblock Motion robust high-speed light-weighted object detection with event camera.
\newblock \emph{{IEEE Trans. Instru. and Measurement}}, 72:\penalty0 1--13, 2023.

\bibitem[Liu et~al.(2021)Liu, Lin, Cao, et~al.]{swins_trans}
Ze Liu, Yutong Lin, Yue Cao, et~al.
\newblock Swin transformer: Hierarchical vision transformer using shifted windows.
\newblock In \emph{{Proc IEEE Int. Conf. on Computer Vision}}, pages 10012--10022, 2021.

\bibitem[Peng et~al.(2024)Peng, Li, Zhang, et~al.]{event_detector_2024_SAST}
Yansong Peng, Hebei Li, Yueyi Zhang, et~al.
\newblock Scene adaptive sparse transformer for event-based object detection.
\newblock In \emph{{Proc. IEEE/CVF Conf. Comput. Vis. Pattern Recognit.}}, 2024.

\bibitem[Perot et~al.(2020)Perot, De~Tournemire, Nitti, et~al.]{event_dataset_2020_NIPS}
Etienne Perot, Pierre De~Tournemire, Davide Nitti, et~al.
\newblock Learning to detect objects with a 1 megapixel event camera.
\newblock \emph{{Proc Annual Conf. on Neural Infor. Pro. systems}}, pages 16639--16652, 2020.

\bibitem[Schaefer et~al.(2022)Schaefer, Gehrig, and Scaramuzza]{event_gnn_2022}
Simon Schaefer, Daniel Gehrig, and Davide Scaramuzza.
\newblock Aegnn: Asynchronous event-based graph neural networks.
\newblock In \emph{{Proc. IEEE/CVF Conf. Comput. Vis. Pattern Recognit.}}, pages 12371--12381, 2022.

\bibitem[Schuster and Paliwal(1997)]{RNN_1997_TSP}
Mike Schuster and Kuldip~K Paliwal.
\newblock Bidirectional recurrent neural networks.
\newblock \emph{{IEEE Trans. Signal Process}}, 45\penalty0 (11):\penalty0 2673--2681, 1997.

\bibitem[Smith et~al.(2022)Smith, Warrington, and Linderman]{smith2022simplified}
Jimmy~TH Smith, Andrew Warrington, and Scott~W Linderman.
\newblock Simplified state space layers for sequence modeling.
\newblock \emph{arXiv preprint arXiv:2208.04933}, 2022.

\bibitem[Smith and Topin(2018)]{onecycle}
Leslie~N Smith and Nicholay Topin.
\newblock Super-convergence: Very fast training of neural networks using large learning rates.
\newblock In \emph{{Proc Int. Conf. on Learn. Repre.}}, 2018.

\bibitem[Sun et~al.(2022)Sun, Sakaridis, Liang, et~al.]{EFNet}
Lei Sun, Christos Sakaridis, Jingyun Liang, et~al.
\newblock Event-based fusion for motion deblurring with cross-modal attention.
\newblock In \emph{{Proc Eur. Conf. on Computer Vision}}, pages 412--428. Springer, 2022.

\bibitem[Tavanaei et~al.(2019)Tavanaei, Ghodrati, Kheradpisheh, Masquelier, and Maida]{tavanaei2019deep}
Amirhossein Tavanaei, Masoud Ghodrati, Saeed~Reza Kheradpisheh, Timoth{\'e}e Masquelier, and Anthony Maida.
\newblock Deep learning in spiking neural networks.
\newblock \emph{Neural networks}, 111:\penalty0 47--63, 2019.

\bibitem[Tomy et~al.(2022)Tomy, Paigwar, Mann, et~al.]{effusion_2022}
Abhishek Tomy, Anshul Paigwar, Khushdeep~S Mann, et~al.
\newblock Fusing event-based and rgb camera for robust object detection in adverse conditions.
\newblock In \emph{{Proc IEEE Int. Conf. on Robot. and Automat.}}, pages 933--939. IEEE, 2022.

\bibitem[Tu et~al.(2022)Tu, Talebi, Zhang, et~al.]{maxvit_2022_ECCV}
Zhengzhong Tu, Hossein Talebi, Han Zhang, et~al.
\newblock Maxvit: Multi-axis vision transformer.
\newblock In \emph{{Proc Eur. Conf. on Computer Vision}}, pages 459--479. Springer, 2022.

\bibitem[Tulyakov et~al.(2019)Tulyakov, Fleuret, Kiefel, et~al.]{event_framing_2019_ICCV}
Stepan Tulyakov, Francois Fleuret, Martin Kiefel, et~al.
\newblock Learning an event sequence embedding for dense event-based deep stereo.
\newblock In \emph{{Proc IEEE Int. Conf. on Computer Vision}}, pages 1527--1537, 2019.

\bibitem[Wang et~al.(2020)Wang, Liao, Wu, et~al.]{CSPNet}
Chien-Yao Wang, Hong-Yuan~Mark Liao, Yueh-Hua Wu, et~al.
\newblock Cspnet: A new backbone that can enhance learning capability of cnn.
\newblock In \emph{{Proc. IEEE/CVF Conf. Comput. Vis. Pattern Recognit. workshops}}, pages 390--391, 2020.

\bibitem[Wang et~al.(2024)Wang, Yu, Yu, et~al.]{wang2024towards}
Xiangyuan Wang, Huai Yu, Lei Yu, et~al.
\newblock Towards robust keypoint detection and tracking: A fusion approach with event-aligned image features.
\newblock \emph{{IEEE Robot. Automat. Lett.}}, 2024.

\bibitem[Woo et~al.(2018)Woo, Park, Lee, and Kweon]{cbam}
Sanghyun Woo, Jongchan Park, Joon-Young Lee, and In~So Kweon.
\newblock Cbam: Convolutional block attention module.
\newblock In \emph{{Proc Eur. Conf. on Computer Vision}}, pages 3--19, 2018.

\bibitem[Wu et~al.(2017)Wu, Zheng, Yu, et~al.]{RGB_infrared}
Ancong Wu, Wei-Shi Zheng, Hong-Xing Yu, et~al.
\newblock Rgb-infrared cross-modality person re-identification.
\newblock In \emph{{Proc. IEEE/CVF Conf. Comput. Vis. Pattern Recognit.}}, pages 5380--5389, 2017.

\bibitem[Yu et~al.(2024)Yu, Li, Yang, et~al.]{huai_evlsd}
Huai Yu, Hao Li, Wen Yang, et~al.
\newblock Detecting line segments in motion-blurred images with events.
\newblock \emph{{IEEE Trans. Pattern Anal. Mach. Intell.}}, 46\penalty0 (5):\penalty0 2866--2881, 2024.

\bibitem[Zhang et~al.(2022)Zhang, Dong, Zhang, et~al.]{event_spike_2022_CVPR}
Jiqing Zhang, Bo Dong, Haiwei Zhang, et~al.
\newblock Spiking transformers for event-based single object tracking.
\newblock In \emph{{Proc. IEEE/CVF Conf. Comput. Vis. Pattern Recognit.}}, pages 8801--8810, 2022.

\bibitem[Zhang et~al.(2023)Zhang, Wang, Liu, et~al.]{event_2023_tracking}
Jiqing Zhang, Yuanchen Wang, Wenxi Liu, et~al.
\newblock Frame-event alignment and fusion network for high frame rate tracking.
\newblock In \emph{{Proc. IEEE/CVF Conf. Comput. Vis. Pattern Recognit.}}, pages 9781--9790, 2023.

\bibitem[Zhou et~al.(2023)Zhou, Wu, Boutteau, et~al.]{cbam_detection}
Zhuyun Zhou, Zongwei Wu, R{\'e}mi Boutteau, et~al.
\newblock Rgb-event fusion for moving object detection in autonomous driving.
\newblock In \emph{{Proc IEEE Int. Conf. on Robot. and Automat.}}, pages 7808--7815. IEEE, 2023.

\bibitem[Zhu et~al.(2019)Zhu, Yuan, Chaney, and Daniilidis]{event_framing_2019_CVPR}
Alex~Zihao Zhu, Liangzhe Yuan, Kenneth Chaney, and Kostas Daniilidis.
\newblock Unsupervised event-based learning of optical flow, depth, and egomotion.
\newblock In \emph{{Proc. IEEE/CVF Conf. Comput. Vis. Pattern Recognit.}}, pages 989--997, 2019.

\bibitem[Zubi\'c et~al.(2024)Zubi\'c, Gehrig, and Scaramuzza]{event_detector_2024_SSM}
Nikola Zubi\'c, Mathias Gehrig, and Davide Scaramuzza.
\newblock State space models for event cameras.
\newblock In \emph{{Proc. IEEE/CVF Conf. Comput. Vis. Pattern Recognit.}}, 2024.

\end{thebibliography}
}


\end{document}